\documentclass[preprint,3p,twocolumn]{elsarticle}
\usepackage{graphicx}
\usepackage{mathrsfs}
\usepackage{algorithm,algorithmicx, algpseudocode}
\usepackage{amsmath}
\usepackage{caption}
\usepackage{multicol}
\usepackage[export]{adjustbox}
\usepackage{lineno,hyperref}
\usepackage{fancyvrb}
\usepackage{bchart}

\modulolinenumbers[5]

\def\CircleArrow{\hbox{$\circ$}\kern-1.5pt\hbox{$\rightarrow$}}

\def\bsq#1{%both single quotes
\lq{#1}\rq}

\algnewcommand\algorithmicnot{\textbf{not}}
\algdef{SE}[IF]{IfNot}{EndIf}[1]{\algorithmicif\ \algorithmicnot\ #1\ \algorithmicthen}{\algorithmicend\ \algorithmicif}%

\journal{Journal of \LaTeX\ Templates}

\let\today\relax
\makeatletter
\def\ps@pprintTitle{%
    \let\@oddhead\@empty
    \let\@evenhead\@empty
    \def\@oddfoot{\footnotesize\itshape
         {~} \hfill\today}%
    \let\@evenfoot\@oddfoot
    }
\makeatother

\begin{document}

\begin{frontmatter}

\title{An Ad-hoc graph node vector embedding algorithm for general knowledge graphs using Kinetica-Graph$^{\vcenter{\hbox{\includegraphics[height=0.35cm]{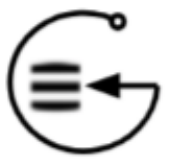}}}\dagger}$}

\author{B. Kaan Karamete\corref{cc}} 
\author{Eli Glaser\corref{}}

\cortext[cc]{\it Corresponding author: Bilge Kaan Karamete, \\ 
kkaramete@kinetica.com, karametebkaan@gmail.com\\ 
$\dagger$Kinetica-Graph: https://arxiv.org/abs/2201.02136 }
\address{Kinetica DB Inc. \break 901 North Glebe Road, Arlington, Virginia 22203}

\begin{abstract}

This paper discusses how to generate general graph node embeddings from knowledge graph representations. The embedded space is composed of a number of sub-features to mimic both local affinity and remote structural relevance. These sub-feature dimensions are defined by several indicators that we speculate to catch nodal similarities, such as hop-based topological patterns, the number of overlapping labels, the transitional probabilities (markov-chain probabilities), and the cluster indices computed by our recursive spectral bisection (RSB) algorithm. These measures are flattened over the one dimensional vector space into their respective sub-component ranges such that the entire set of vector similarity functions could be used for finding similar nodes. The error is defined by the sum of pairwise square differences across a randomly selected sample of graph nodes between the assumed embeddings and the ground truth estimates as our novel loss function defined by Equation~\ref{eqn:loss}. The ground truth is estimated to be a combination of pairwise Jaccard similarity and the number of overlapping labels. Finally, we demonstrate a multi-variate stochastic gradient descent (SGD) algorithm to compute the weighing factors among sub-vector spaces to minimize the average error using a random sampling logic. 

\noindent 

\end{abstract}


\end{frontmatter}

\section{Introduction}

There is not a definitive way to represent fixed dimension vector node embed-dings from variable dimension general knowledge graph connections (relations as edges). The simple reasoning is that there is really no rule in a general graph connection sense. Nevertheless, in the last decade, researchers have been pushing the envelope to apply the success of vector embedding advancements in Large Language Models (LLM) over to the general context knowledge graphs. A number of novel algorithms, namely, node2vec and word2vec~\cite{node2vec,word2vec} devised with lots of success in language semantics and undoubtedly opened doors for today's many AI applications that seem to be championed as the master key solution for most of our engineering problems, if not all (exclusions are mostly in multi constraint optimization problems, such as supply chain logistics and optimal fleet routes). Though, possibly a more humble acceptance of LLM's superiority is when there is supposedly a hidden pattern among word pairings that can be put in a neural net machinery to minimize the error between the assumed solution and the known ground truth based on either some training data (supervised) or a logic of differentiation (unsupervised/reinforced) ~\cite{machinelearning}.

However different in its details of minimizing errors to create these sophisticated language models that would produce intended outcomes, its superiority mainly lies in the deterministic nature of the input and the output with the additions of some fuzziness so that near-reality outcomes could be achieved~\cite{llm1,llm2}. In a general knowledge graph sense, however, there is no language ruling for how a node \bsq{Tom} is connected to \bsq{Bill} or \bsq{Jane} or to the country of his/her birth place, certainly none better than mere connections as node to node \bsq{relation}s. It is always possible however, the problem can be cast into a language model by connecting these general nodal relations around building sentences and paragraphs. These embeddings are broadly generalized into \bsq{translational} and \bsq{semantic} categories and extensively surveyed in ~\cite{kgembed1, kgembed2} where the former is distance based and the latter is relation centric. 

The intention of this paper is not to find this mapping from unstructured knowledge graphs to language graph models so we could apply the celebrated node2vec method to create nodal embed-dings for similarity analysis. Our goal, however, is to create an ad-hoc mapping framework that is based on each graph's own analytics and using as much machinery as possible from LLM technology to mimic similarities between the node pairs and combine \bsq{translational} and \bsq{semantic} mappings together. Perhaps, one could rephrase that a graph is its own AI model where its connections reveal the unstructured information in its most true form and any other representation is just an approximation at best. 

In this spirit,  we try to come up with a vector embedding in which we use a number of graph predicates, such as topological hop-patterns,  common labels, transitional probabilities via Markov-chain (MC) probabilities, and clustering indices via the recursive spectral bisection (RSB) solver~\cite{rsb} using Kinetica-Graph~\cite{kineticagraph,kineticalabels,kineticarouting,kineticamapmatching}. We would refer these as sub-features and explain each group in Section~\ref{Section:subfeatures} in detail. We then describe a flattening procedure to spread these sub-feature predicates onto the sub-ranges of the vector embedding space in Section~\ref{Section:flattening}. A novel loss function definition is described in Section~\ref{Section:loss} where an average embedding error is assumed to be the sum of square differences between the inner product of nodal vector pairings and pairwise sum of Jaccard scores combined with pairwise common labels. Finally, we will show a stochastic gradient descent (SGD) algorithm~\cite{sgd} that minimizes this average error by adjusting the weights among sub-feature groups in the embedded space in Section~\ref{Section:sgd}. 

\section{Sub-vector features}
\label{Section:subfeatures}

\begin{figure*}
\centering
    \includegraphics[width=\linewidth, keepaspectratio]{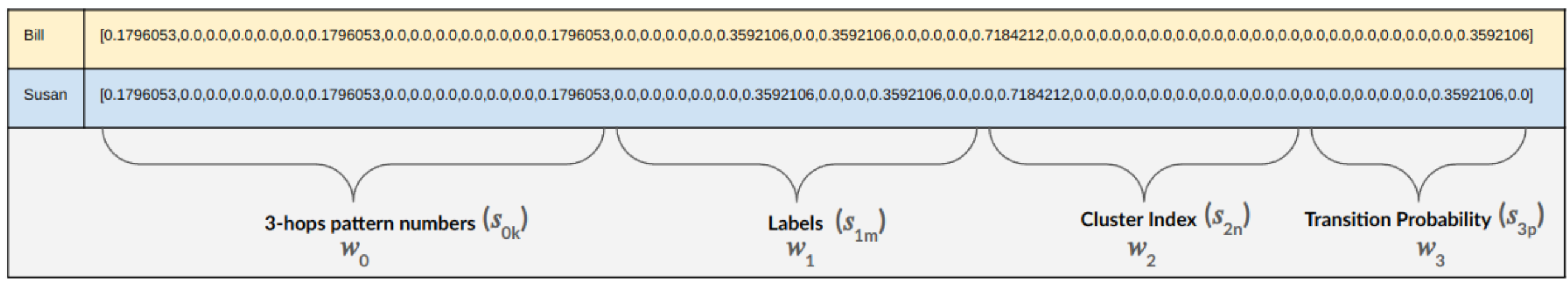}
    \caption{The layout of the sub-features within the vector embedding space; The sub-features, $s_{0..3,k..p}$ per node are hop based topology pattern, associated labels, the cluster index computed using the recursive spectral bisection (RSB) algorithm and the transitional probabilities (markov chain solver), respectively. The $k,m,n,p$ are vector ranges per each sub-feature. Four weight factors, $w_{0..3}$ that will be used to minimize the average total loss per node are eventually multiplied with each value within the sub-range of its respective feature $s_{0..3}$ for the final embedded vector content per graph node.}
    \label{Figure:features}
\end{figure*}

The vector space is divided into sub-group range of indices that are indicative of ad-hoc graph predicates as shown in Figure~\ref{Figure:features}. This is crucial since a value at an index location would have a specific meaning for every node and a share in similarity score when it is inner product-ed with that of another node. These predicates are specifically chosen to capture the local and remote affinities. The following predicates are chosen per graph node:
\begin{itemize}[-]
\item Hop-patterns 
\item Label index associations
\item Cluster index 
\item Transitional probability
\end{itemize}

These predicates are explained in detail below. 

\subsection{hop-patterns}
\label{Section:hop-patterns}
\begin{figure}
\centering
    \includegraphics[width=\linewidth, keepaspectratio]{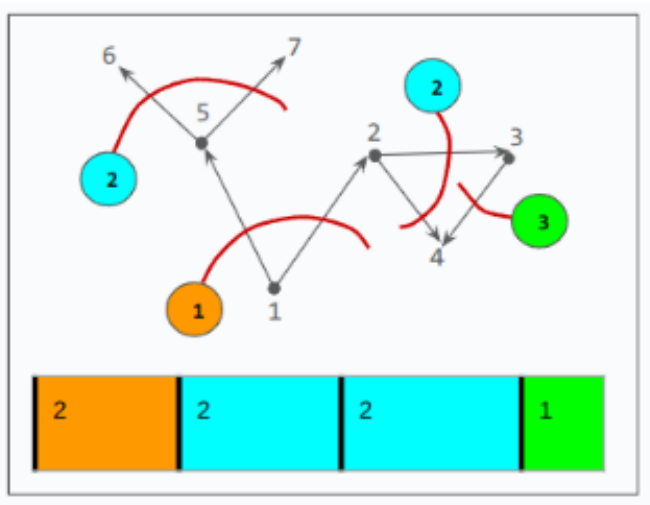}
    \caption{Hop pattern of a node is defined by the number of forks and the number of nodes in each fork arm as shown with the respective colors per hop; e.g., second hop depicted as cyan has two forks with two nodes at each fork arm as shown in the array below.}
    \label{Figure:hoppatterns}
\end{figure}

The first feature predicate encompasses a range of indices to depict hop based pattern numbers as shown in Figure~\ref{Figure:hoppatterns}. Hop pattern of a node is defined by the number of forks and the number of nodes in each fork arm as shown with the respective colors per hop; e.g., second hop depicted as cyan has two forks with two nodes at each fork arm. This is not full-fledged topological pattern matching, since that would require the node indices instead of the number of nodes at breadth-first search (bfs) adjacency traversal. The reason why we can not use the node indices in the vector is that it does not have a meaning as a value subject to inner product. If we can find a better means to universally reflect node indices in vector embeddings, this sub-feature could be replaced with much accurate values, but for now, we'll be using this light weight topological feature. Maximum number of hops is added as an option to the embedding algorithm as the set of pattern based numbers can slide within the array based on this option.

\subsection{Label indices}
\label{Section:labelindices}

We have devised in~\cite{kineticalabels} an efficient mechanism to attach multiple labels to nodes and edges. The labels are stored with their unique indexes in the graph db. This feature has as many sub-range indices in the vector as there are unique labels in the graph (that has node-associations). The idea as similar to hop patterns is for these array indexes to have an absolute meaning throughout the nodes, i.e., if a label index is common to a number of nodes, the specific array index for that label should be turned on for all those nodes that share the same label. The label indices are depicted in Figure~\ref{Figure:features} as $k,m,n,p$ for each sub-feature, respectively.

\begin{figure}
\centering
    \framebox{\includegraphics[width=0.9\linewidth, keepaspectratio]{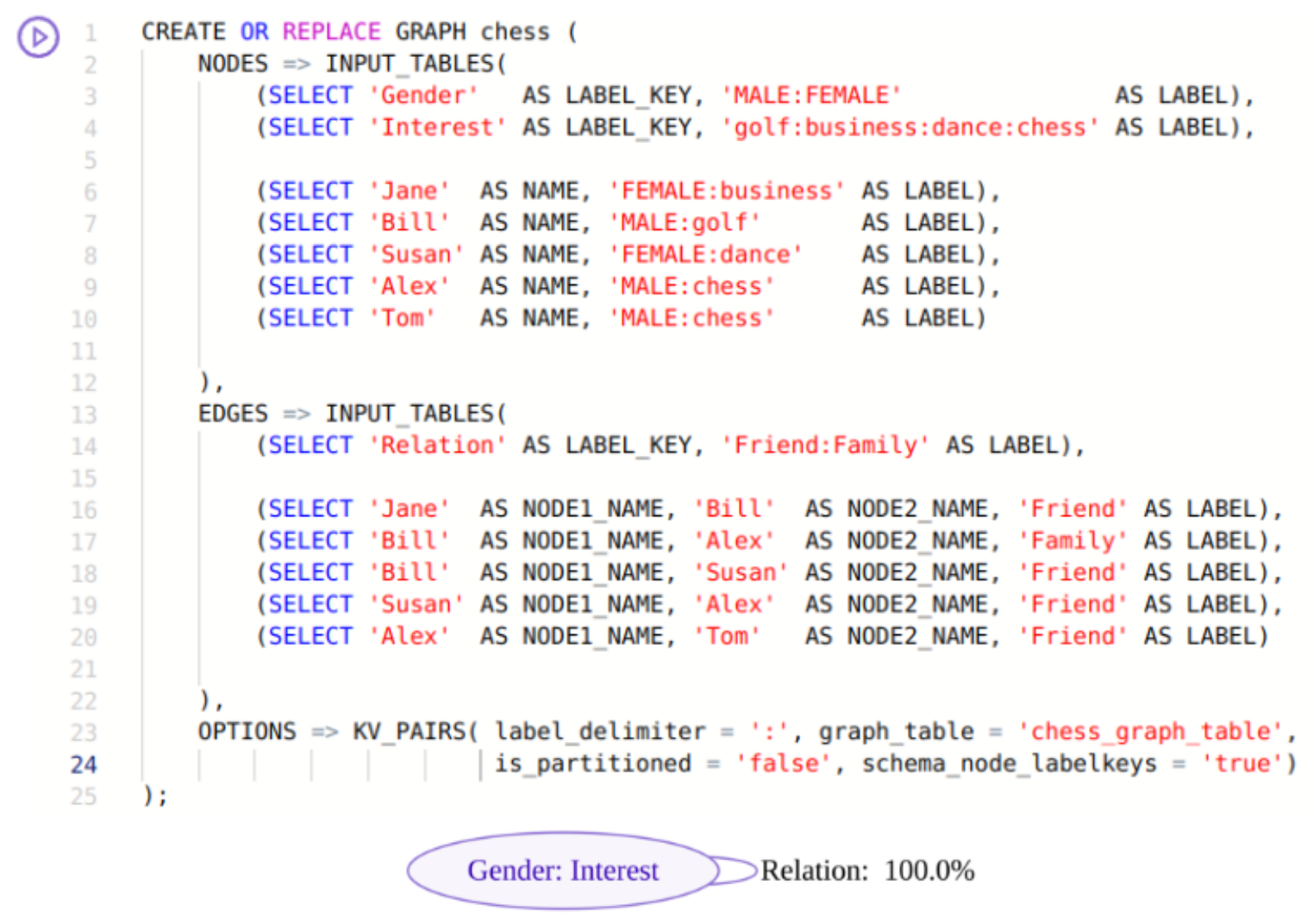}}
    \caption{Graph-SQL syntax for Kinetica-Graph's create/graph Restful API. The nodes and edges components are depicted explicitly, i.e., with constants instead of reading from table columns (unless the example is simple the usual way is to list the nodes/edges in a DB table or stream in). Bottom is showing  the chess graph ontology using the label keys; all edges ($100 \%$) are in between \textit{Gender} and \textit{Interest} labeled nodes via \textit{Relation} group edge label key.}
    \label{Figure:creategraph}
\end{figure}

\begin{figure}
\centering
    \includegraphics[width=0.7\linewidth, keepaspectratio]{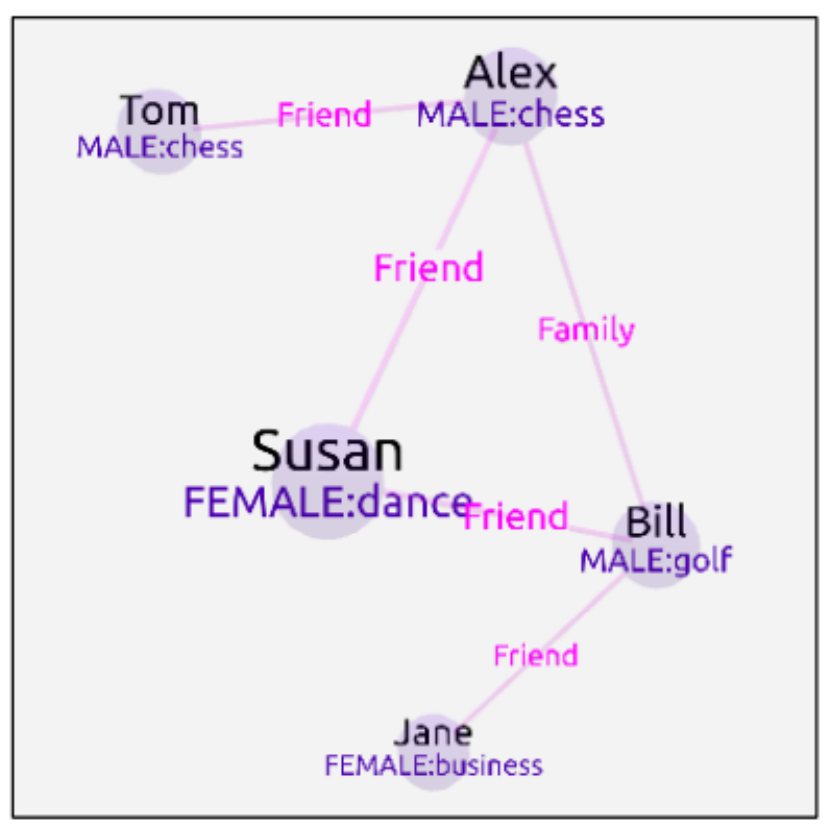}
    \caption{3D visualization of the graph generated by tha call in~Figure\ref{Figure:creategraph} with node/edge label associations using d3's force layout}
    \label{Figure:graphschema}
\end{figure}

\begin{figure}
\centering
    \includegraphics[width=0.6\linewidth, keepaspectratio]{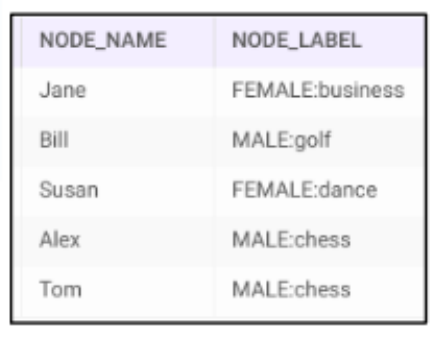}
    \caption{The response of Kinetica-Graph's create/graph call depicting node-label associations as a relational DB table. E.g.: \textit{Alex} and \textit{Tom} has two common labels, namely, \textit{chess} and \textit{MALE}.}
    \label{Figure:commonlabels}
\end{figure}

To illustrate label associations and indices, the generation of a simple wiki-graph is depicted with the SQL syntax of Kinetica-Graph's restful API in Figure~\ref{Figure:creategraph}. Node labels are listed in the response of the create-graph call in Figure~\ref{Figure:commonlabels} as a DB table. A 3D visualization of the graph with nodal label associations are also shown in Figure~\ref{Figure:graphschema}.

\subsection{Cluster indices}
\label{Section:clusterindices}

\begin{figure*}
\centering
    \includegraphics[width=0.7\linewidth, keepaspectratio]{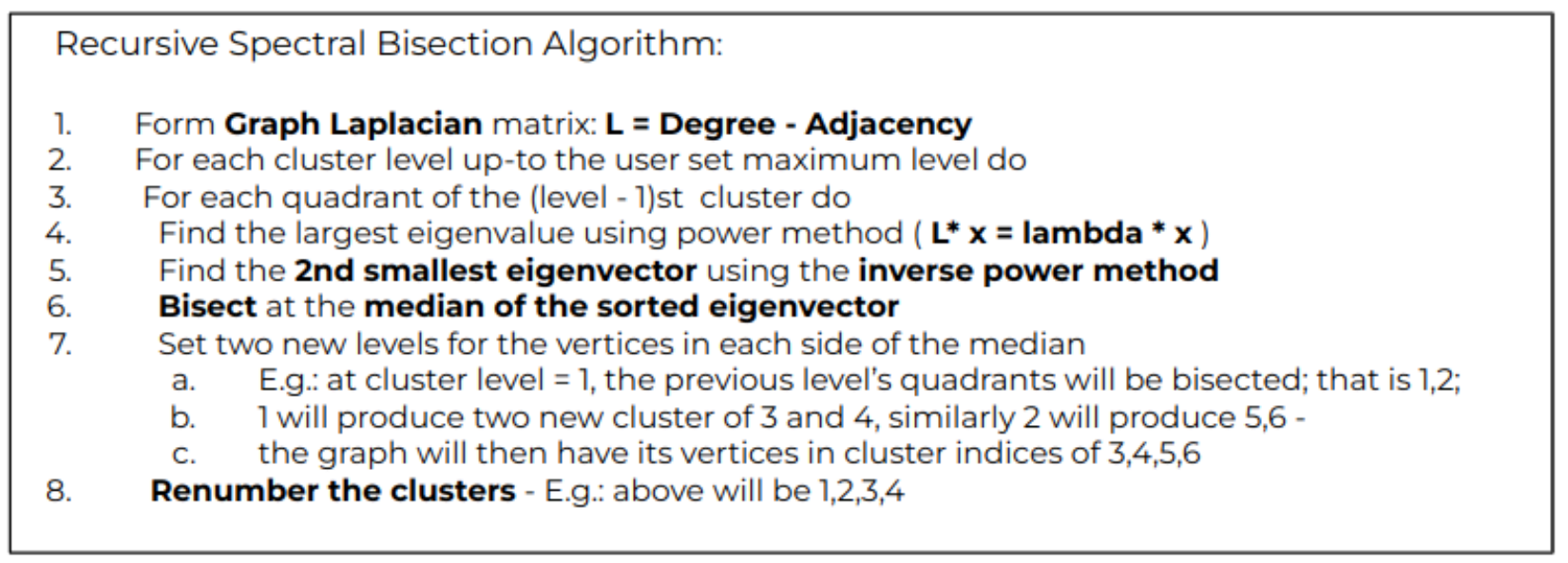}
    \caption{Steps of Recursive Spectral Bisection (RSB) Algorithm}
    \label{Figure:rsb}
\end{figure*}

\begin{figure}
\centering
    \includegraphics[width=\linewidth, keepaspectratio]{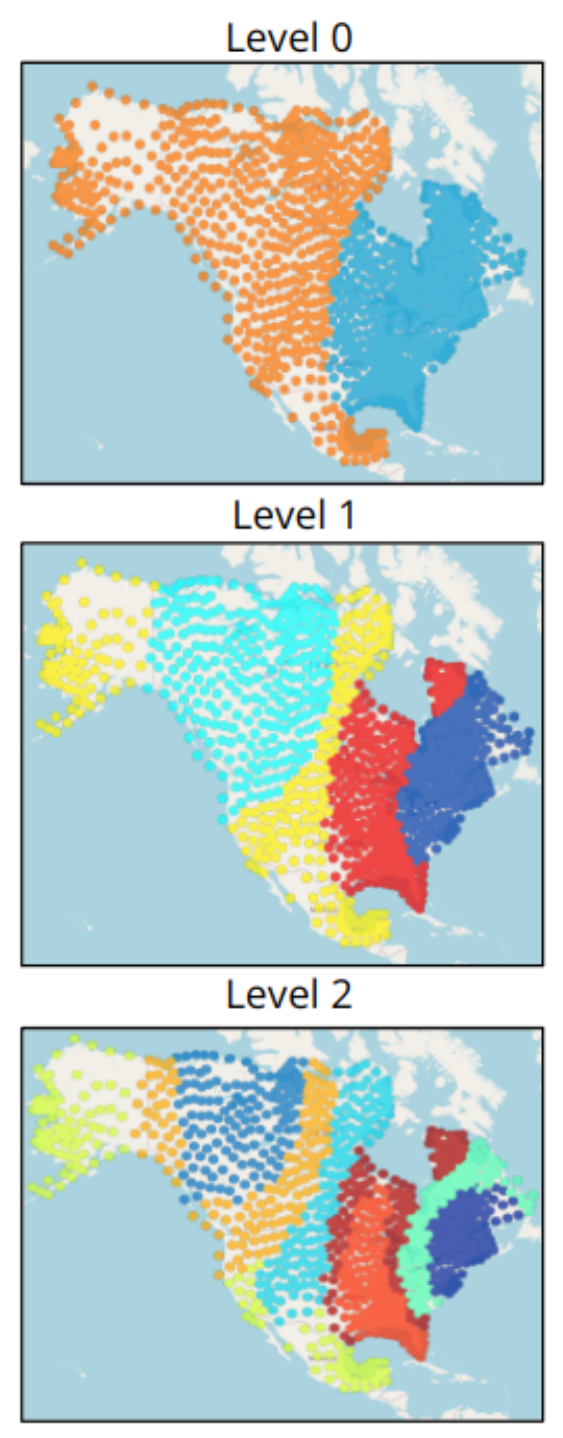}
    \caption{The application of the RSB algorithm over the mesh graph of the continental USA. The three levels of bisection creates $8$ (default number of clusters) clusters from top to bottom, respectively.}
    \label{Figure:rsbusa}
\end{figure}

\begin{figure}
\centering
    \includegraphics[width=\linewidth, keepaspectratio]{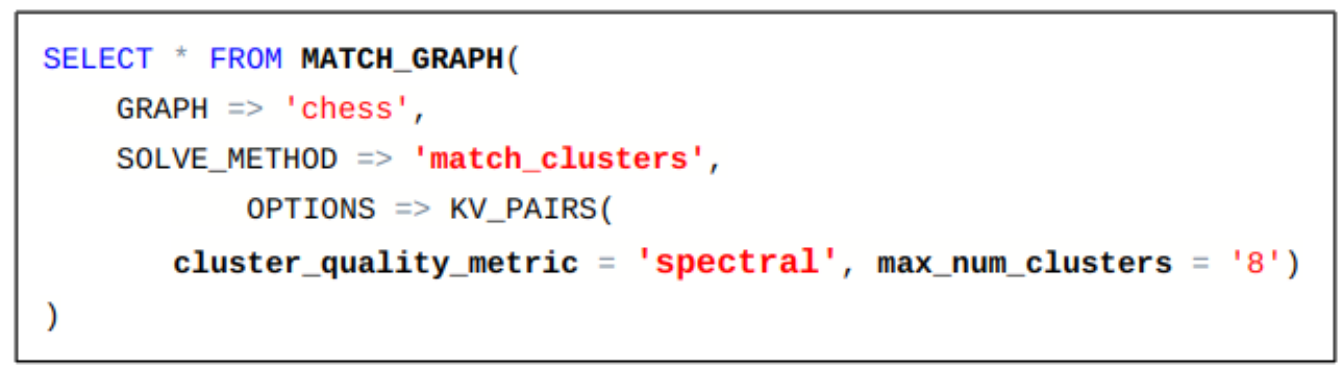}
    \caption{The Graph-SQL syntax of match/graph restful API to generate clusters using RSB method. Another option is the \textit{Louvain} clustering with more resource requirements.}
    \label{Figure:matchclusters}
\end{figure}

A recursive spectral bisection algorithm is used to compute the cluster indexes of graph nodes. The idea in this Kinetica-Graph implementation is inspired from~\cite{rsb} where the sorted second smallest eigenvector of the graph Laplacian is used in bisecting at the median location at each recursive split as depicted in Algorithm~\ref{Figure:rsb}. The choice of partitioning for the endpoint match/graph with \bsq{spectral} option is particularly preferred for its speed of execution and low resource allocation requirement compared to the \textit{Louvain} clustering~\cite{louvain} (another option in the solver) as shown in the Figure~\ref{Figure:matchclusters}. A geometrical example of the RSB method with three levels of bisections can be seen in Figure~\ref{Figure:rsbusa}. 
  
\begin{figure}
\centering
    \includegraphics[width=0.4\linewidth, keepaspectratio]{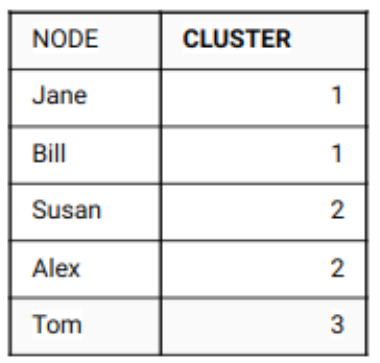}
    \caption{The output of the RSB clustering - the cluster indexes per node as a DB table for the simple wiki-graph example}
    \label{Figure:clusters}
\end{figure}

The cluster index per node is pushed into the respective sub-range allocated for this feature. The width of this feature over the vector space can be scaled by overriding the default value of $8$ via the maximum number of clusters option to the embedding solver. The output of the RSB clustering is shown as a DB table in Figure~\ref{Figure:clusters}. 

\subsection{Transitional Probabilities}
\label{Section:clusterindices}

Inspired from the Pagerank algorithm~\cite{pagerank}, our novel probability ranking solver uses the same equation depicted in~\ref{eqn:transition} with a modified transition probability flux $p_{ij}$ where it is computed to be the ratio of incoming adjacent edge weights (connecting nodes $i$ and $j$) within the immediate neighbor $B(i)$ to each node $i$. These nodal scalar $p_i$ values are iterated at each traversal loop converging to a steady state where the maximal change in $p_i$ is less than a small threshold. The ranking factor $r$ is assumed to be $0.15$ so that every node will have a small amount of uniform probability ($r$ divided by the number of graph nodes $NV$) to account particularly for nodes with no incoming edges. 

\begin{align}
\label{eqn:transition}
p_i = (1-r)\sum_{j\in B(i)}p_{ij} + \frac{r}{NV}
\end{align}

\begin{figure}
\centering
    \includegraphics[width=0.7\linewidth, keepaspectratio]{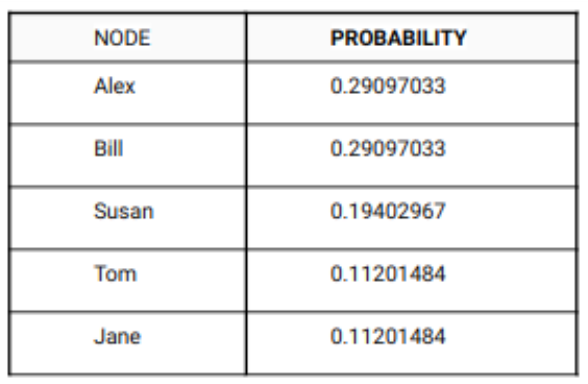}
    \caption{The results of the probability ranking solver per node as a DB table - the probabilities are scaled between a zero (small value) and one, with the sum equal to one.}     
    \label{Figure:markovprob}
\end{figure}

These transitional probabilities are scaled between zero and one and the sub-range for this feature over the vector is allocated to a preset division. For example, if a particular node's probability is $0.25$, with the default range of ten, the third index within the sub-vector range is turned on. It is also worth mentioning that this solver behaves exactly like a Pagerank solver for the uniform identical edge weights scenario where the transitional probability defined above simply becomes the incoming valence rank.

\begin{figure}
\centering
    \includegraphics[width=0.8\linewidth, keepaspectratio]{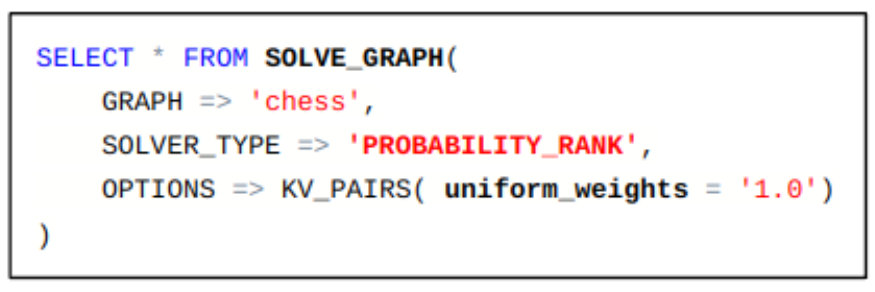}
    \caption{Graph-SQL syntax in table function form of Kinetica-Graph's solve/graph restful API with uniform weights of $1$. Equal weight distribution of the solver mimics the Pagerank results.}
    \label{Figure:matchmarkov}
\end{figure}

\section{Flattening}
\label{Section:flattening}
Hop-pattern numbers are per each fork of a hop. Therefore, primary flattening occurs in mapping this two dimensional information over the one dimensional vector space. Furthermore, a secondary flattening happens for laying the sub-feature's own indexing after the previous feature's flattened index location. Label and cluster indices do not require flattening but their respective vector locations have to be shifted after the prior feature's index range. Transitional probability, however, would require to map the continuous scalar probability values to be bucketed over a preset number of interval indices. This is easily accomplished by partitioning the unit range equally over a pre-set number of intervals (can be modified by the user). The vector size formula is given in Equation~\ref{eqn:flatten} where each sub-feature's respective ranges are summed. The parameters \textit{max\_num\_clusters}, \textit{max\_hops} are user driven and \textit{max\_forks\_perhop}, \textit{max\_edges\_perfork} are set implicitly to minimize the number of pattern indices within the overall vector dimension.

\begin{align}
\label{eqn:flatten}
Vector\_size=&~max\_patterns~+ \nonumber \\
 &~max\_labels~+ \nonumber \\
 &~max\_num\_clusters~+ \nonumber \\
 &~num\_probabilities \nonumber \\
 where& \nonumber \\ 
 max\_patterns=&~max\_hops~\times \nonumber \\
               &~max\_forks\_perhop~\times \nonumber \\
               &~max\_edges\_perfork 
\end{align}

Before the normalization process, these vector values within each sub-feature are multiplied by a feature specific weight parameter. We can extend the number of sub-features in our embedding framework, however, at the time of writing this manuscript, we are currently having four sub-features, hence, we have only four weight parameters that we will use for the purpose of finding out which of these sub-features are intrinsically more dominant over each other for the specific graph in contention using an equal error distribution and  minimization procedure that will be explained in Section~\ref{Section:loss} and~\ref{Section:sgd}.

Having multiplied by their respective weights, the vectors are then normalized so that their inner products can be used for vector similarity analysis. The unit vector embedding values for the node pair \bsq{Bill} and \bsq{Susan} for the simple wiki-graph is shown in Figure~\ref{Figure:features}.

\subsection{Quantizing}
\label{Section:quantizing}

\begin{figure*}
\centering
    \includegraphics[width=0.7\linewidth, keepaspectratio]{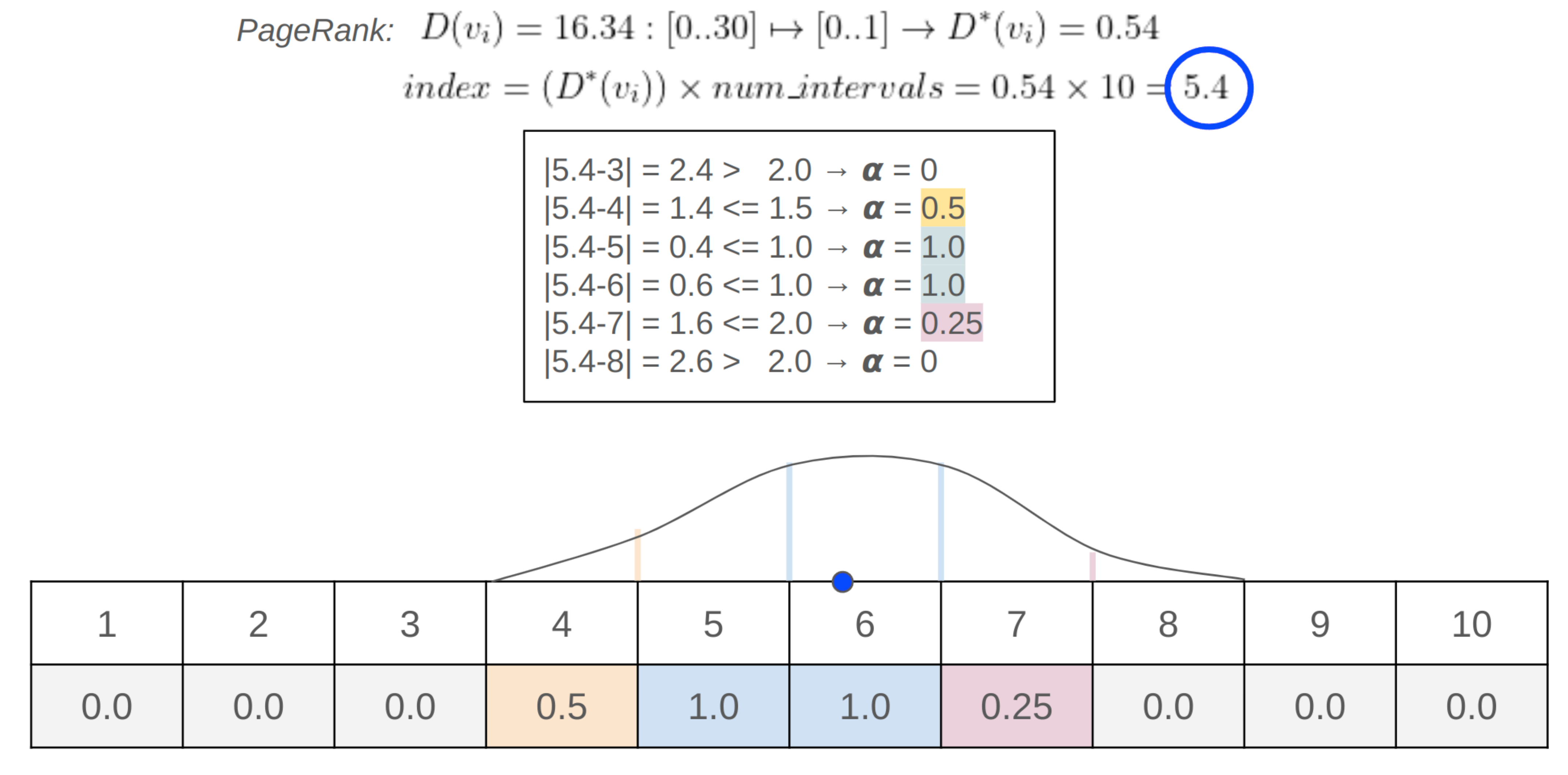}
    \caption{Quantizing: piece-wise dispersion of the float value of $5.4$ for the pagerank score to a number of nearby vector indexes based on the deviations from the exact value. Spread of nodal values mapped between zero and one using a sigmoid is followed by the quantizing step that increase the chance of capturing more similarities among node embeddings.}
    \label{Figure:quantizing}
\end{figure*}

Straightforward mapping of the float predicates over the vector indices will only provide inner products for those whose predicates fall onto the exact vector index. However, this is not realistic since the integer index equivalent for a float value corresponding to a pagerank or distance (extended version of the embedding solver includes distances) score is too restrictive and would certainly miss close but off-the-index values in similarity (inner-product) computations. For instance, a value of $5.4$ should not just turn on the $5^{th}$ index (in a $10$ slot sub-range dedicated for the predicate) but the nearby indices based on its deviation from the exact index, in a hat-like diffusing behavior. Hence, we have developed a \textit{quantizing} logic to diffuse these values over a range of nearby indices as shown in Figure~\ref{Figure:quantizing} where in the particular example not only the $5^{th}$ but the nearby $4^{th}$, $6^{th}$ and $7^{th}$ indices had the effect of dispersion based on the  deviations of these indices from the exact value in a piece-wise fashion. \textit{Quantizing} is a key concept that helps increase the chance of capturing potential similarities among the node embeddings. 

\section{Loss Function}
\label{Section:loss}

The concept of assuming the total error distributed evenly across the nodal pairs is inspired from the computational mechanics field~\cite{zsquare}, specifically in finite element analysis defined as \textit{z-square} where the elemental errors are aggregated over the entire domain and then divided equally over the finite elements. Similarly, in our sub-feature weight optimization, we can define the total error as the aggregated sum of the differences between the inner product of each node against every other in a subset of the graph and the ground truth estimates for each pair. Specifically, the error is defined by the sum of pairwise square differences across a randomly selected sample of graph nodes between the assumed embeddings and the ground truth estimates as our novel loss function defined by Equation~\ref{eqn:loss}. The ground truth is estimated to be a combination of pairwise Jaccard similarity and the number of overlapping labels. 

%\begin{minipage}{\textwidth}

\begin{table*}[!ht]
\large
\begin{subequations}
\label{eqn:loss}
\begin{align}
Loss_{i} &= \frac{1}{nk}\sum_k^{nk}{\left \|  <f_i,f_k> - \left ( \alpha~jac(i,k) + (1-\alpha) \frac{labels(i)\cap 
 labels(k)}{\sum_{k}{distinct(labels(k))}} \right ) \right \|_2} 
  \\ \nonumber \\
<f_i,f_k> &= w^i_r w^k_r s^i_{rj} s^k_{rj}  
  \\ \nonumber \\
w_j \leftarrow &~min \left (   \frac{ \sum_i(Loss_i)} {NV} \right ) ~~\forall{j} ~ \mid ~ j=1..4 
\end{align}
\end{subequations}
\end{table*}
%\end{minipage}
%\clearpage

Loss function is defined per node $i$ such that the goal is to find the average difference aggregated over all the pairs from the node $i$ to all other $nk$ number of nodes. The similarity, i.e., the inner product between the vector embeddings of $f_i$ and $f_k$ is subtracted from the pairwise sum of Jaccard similarity score and the number of overlapping labels between the pairs as our revised ground truth estimate as shown in Equation~\ref{eqn:loss}(a). The $\alpha$ value is  chosen to be $0.5$ which is the mean of the two measures. The pairwise-error functions is made \textit{L2} norm so that the derivative of the loss function with respect to the four weight parameters would have the truth estimate terms for the optimization algorithm. The vector embedding $f_i$ or $f_k$ is a function of $w_j$'s. The problem is then reduced to applying the minimization procedure to the average of the total sum of the nodal losses across the sub-graph as shown in the Equation~\ref{eqn:loss}(c). 

The selection of the sub-graph is done randomly such that we only grab equal number of graph nodes as batches within each cluster index (computed at the sub-feature creation) so that the random set is representative enough of the entire graph behavior. The total number of the random sampling process is a user defined parameter and usually much less than the original graph size to minimize the overall computational time. 

Finally, by using this random sampling logic a multi-variate stochastic gradient descent (SGD) algorithm is devised to compute the weighing factors in minimizing the average nodal error of Equation~\ref{eqn:loss}(c) in Section~\ref{Section:sgd}.

\begin{figure}
\centering
    \framebox{\includegraphics[width=\linewidth, keepaspectratio]{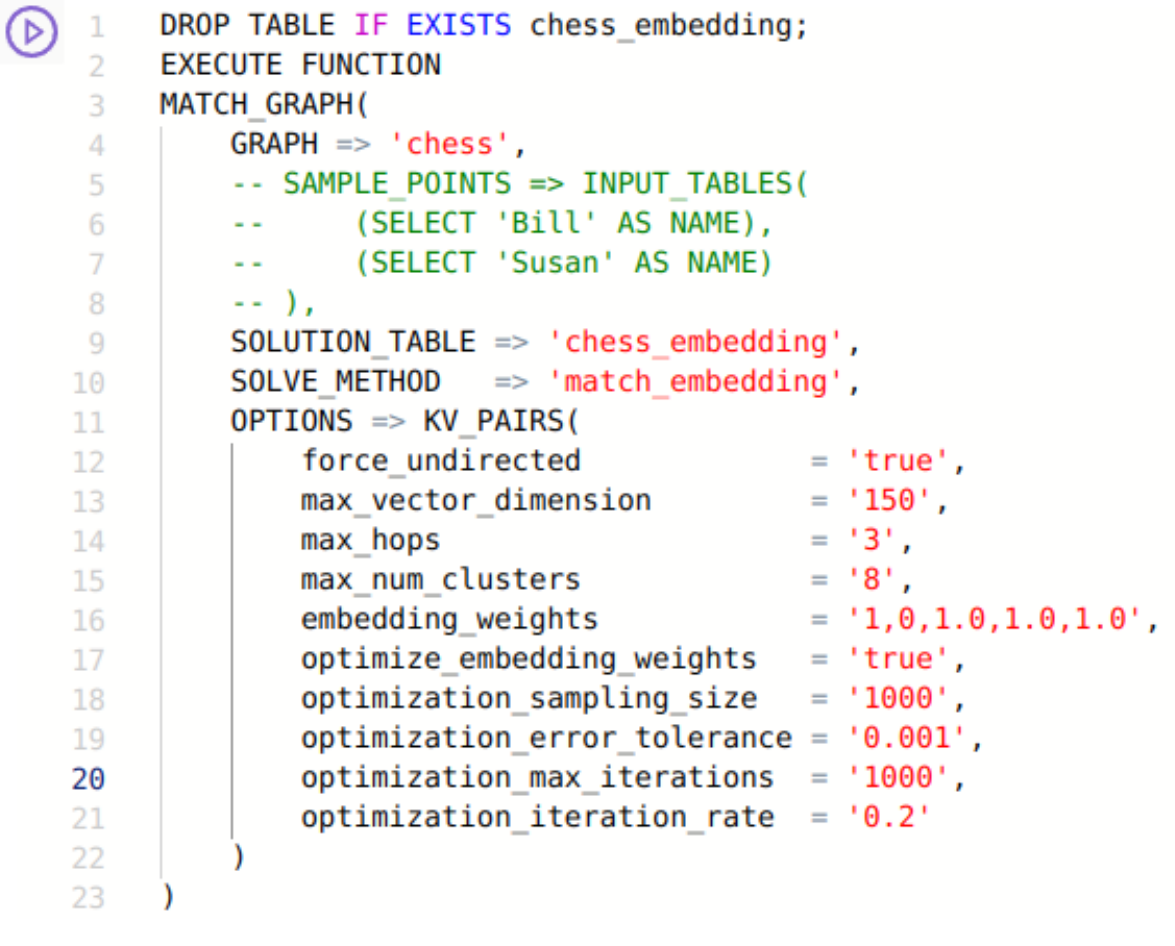}}
    \caption{Graph-SQL syntax of Kinetica-Graph's match/graph restful API with the embedding solver and options that shifts the sub-ranges of the features in the embedded vector space. If the optional \textit{SAMPLE\_POINTS} component is not commented, the solver would output only the pairs specified via the component's combo two-tuple identifiers.}
    \label{Figure:matchembedding}
\end{figure}

\section{Stochastic Gradient Descent}
\label{Section:sgd}

The sum of pairwise differences between the inner product of pairs for node $i$, node $k$ and the ad-hoc ground truth from Jaccard scores with overlapping label count ratio as depicted in Equation~\ref{eqn:loss}(a) is dependent on the unknown terms $w_j$ as the weight factor of each four sub-features. The selection of the set of graph nodes where each node is paired with every other in the set is important in finding these optimal weights. The process needs to include nodes to have a good representation of the entire graph behavior. We have opted to sample this set randomly (stochastic) with a caveat of picking the batch of nodes from each cluster index group where we have already computed in constructing the sub-features. The number of nodes in this random selection process is user specified, however, it needs to be much less than the original graph size so that the SGD iterations would not be prohibitive. 

\begin{figure*}
\centering
    \framebox{\includegraphics[width=0.75\linewidth, keepaspectratio]{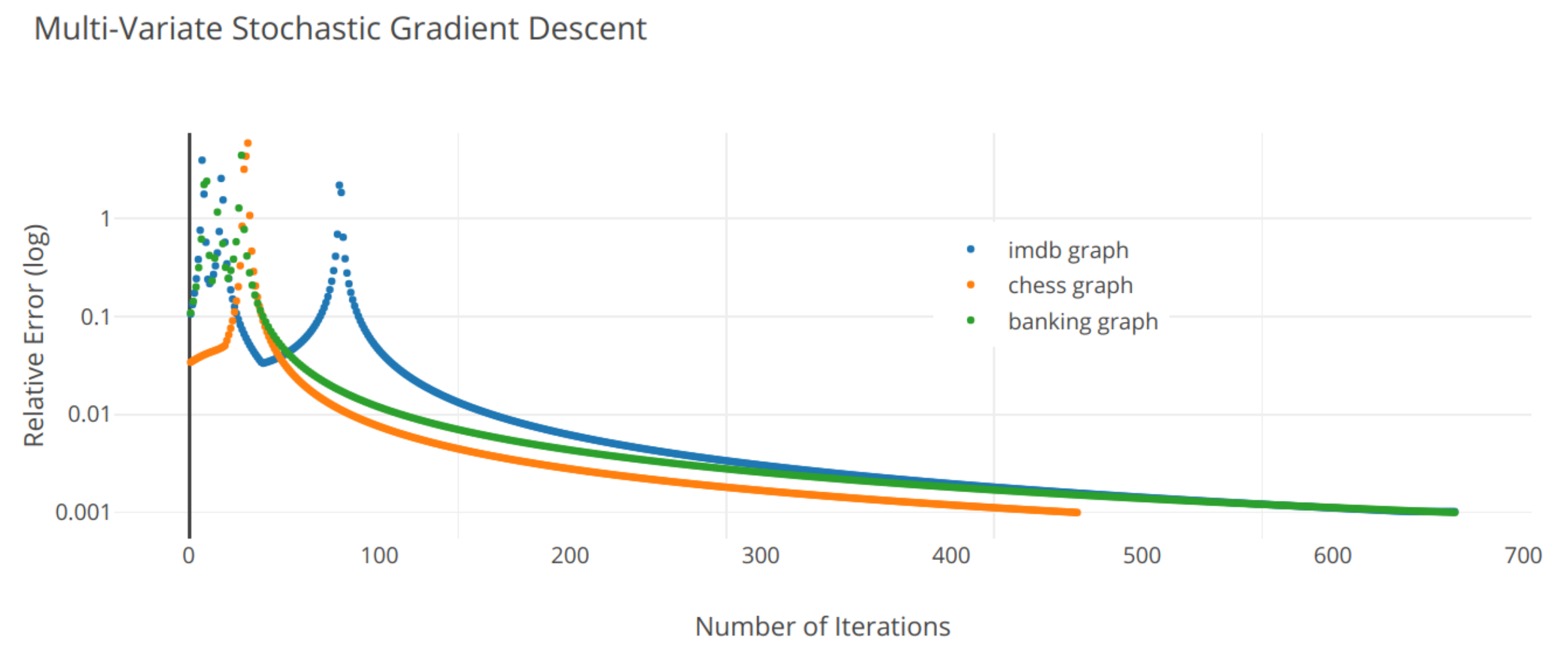}}
    \caption{Stochastic Gradient Descent Convergence}
    \label{Figure:sgd}
\end{figure*}

The next step is taking the derivative of the nodal average of the loss per each of these weight parameters and move against the direction of the gradient of each weight to minimize the loss. The incremental update on each unknown weigth is immediately made to reflect its impact on the next unknown weight variable computation as shown in Equation~\ref{eqn:sgd} and~\ref{eqn:sgdder} in which $w_j^{(k+1)}$ is the next $(k+1)^{th}$ iteration on the $j^{th}$ weight parameter. This approach is not a guaranteed outcome in accelerating the convergence. Other alternative approaches such as \textit{batching} or \textit{mini-batching} discussed extensively in~\cite{sgd} might provide better computational outcome. However, we think this is beyond the scope of this study and our findings are satisfactory computationally for the cases we have tried so far. The iterations are continued if the relative incremental iteration delta of all unknown weights go below a preset user threshold (default value is 0.001) or the number of epoch iterations reaches the upper limit (default is 100) which is also a user prescribed parameter of the solver as shown in Figure~\ref{Figure:matchembedding}. The convergence history plot between the number of iterations and the error is also shown in Figure~\ref{Figure:sgd} for the knowledge graphs of varying sizes from a few hundred to 100 million nodes. The trend in all cases is with early unstable fluctuations and rapid descending to the optimal as expected. The initial weight values and the rate of iteration, also known as training or learning rate, namely, $\beta$ as shown in Equation~\ref{eqn:sgd}, can both also be overridden by the user.

\normalsize{
\begin{align}
\label{eqn:sgd}
w_j^{(k+1)} \leftarrow w_j^k - \left. \beta  \frac{\partial \sum\limits_{i}(Loss_i) \mathbin{/}NV} {\partial w_j} \right|_k
\end{align}
}

\normalsize{
\begin{align}
\label{eqn:sgdder1}
K_{i,m} =  
<f_i,f_m> - \biggl(\alpha~jac(i,m) + \\ \nonumber
(1-\alpha) \frac{labels(i)\cap labels(m)}{\sum_{m}{distinct(labels(m))}} \biggl)
\end{align}
}

\normalsize{
\begin{align}
\label{eqn:sgdder}
\frac{\partial \sum\limits_{i}(Loss_i)}{\partial w_j} = 4\sum\limits_{i}\sum\limits_{m}(K_{i,m} w_j s^i_{j} s^m_{j})
\end{align}
}

\section{Discussion and Conclusions}
\label{Section:conclusions}

We have tried to map unlimited dimension general knowledge graph topology onto a $1-$dimensional vector embeddings by constructing the vector space from features that we think would best resemble local affinity and remote structure so that any vector similarity (inner product) between a node pair would result in the same similarity behavior if we had computed the Jaccard score with the number of common labels between the two nodes of every pair. To this end, the sub-features are chosen to be the predicates of hop-pattern numbers, cluster indices (computed by the recursive spectral bisection (RSB)), associated label indices, and the transitional probability (or the Pagerank score if weights are uniform) as explained in Section~\ref{Section:subfeatures} above. The impact of the vector component sub-features on the similarity can be found by adding a weight parameter to multiply  each of the sub-feature elements within the respective vector sub-range and test the result against an estimated ground truth as explained in Section~\ref{Section:loss}. We have formulated the difference between the inner product of the assumed embeddings and the combined common-labels and Jaccard score as the ground truth, as our ad-hoc \textit{Loss} function as shown in Equation~\ref{eqn:loss}. We then tried to optimize the nodal average of the total loss by applying a stochastic gradient descent (SGD) algorithm to find the unknown weights so that this total average loss is minimized as explained in Section~\ref{Section:sgd} and Equations~\ref{eqn:sgd} and~\ref{eqn:sgdder}.

\begin{figure*}
\centering
    \includegraphics[width=\linewidth, keepaspectratio]{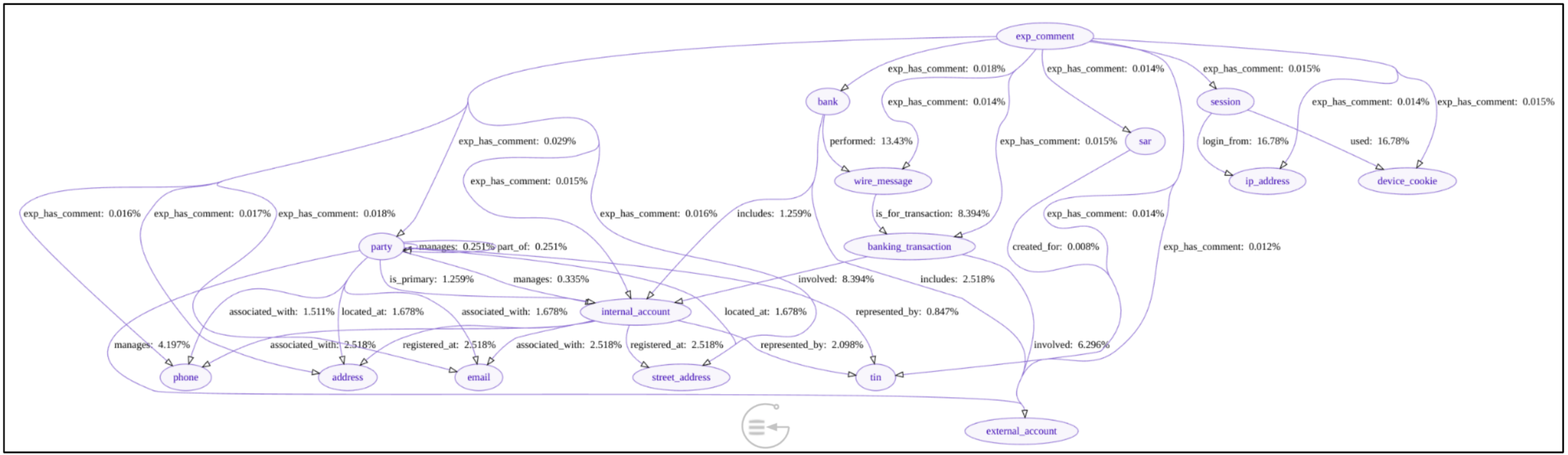}
    \caption{Banking graph ontology with $34$ edge and $16$ node labels with 10 million+ nodes. The percentages show how many actual graph edges are connected between each labels.}    
    \label{Figure:banking}
\end{figure*}

Stochastic process is the selection of the random nodes that will be used in SGD to find the unknown weights. We have chosen these random nodes from each graph cluster in equally numbers.  We use this smaller sub-set of nodes in computing the unknown weights. The assumption of picking this narrow set of nodes from each cluster is to increase the likelihood of better representation of the entire graph since SGD on the entire graph is computationally prohibitive. SGD converges very similarly in our testing of many graphs as shown in Figure~\ref{Figure:sgd}. The banking graph shown, is in 10+ million range (4+ billion case is also used), and its ontology is depicted in Figure~\ref{Figure:banking}. 

\begin{figure}
\centering
    \framebox{\includegraphics[width=0.8\linewidth, keepaspectratio]{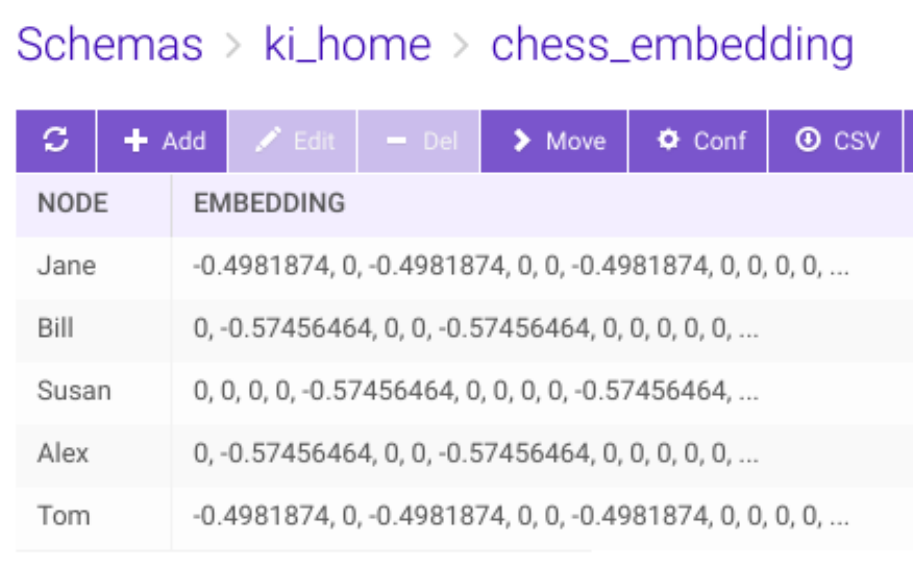}}
    \caption{The vector embedding output table \textit{chess\_embedding} as the result of the Kinetica-Graph embedding solver depicted in Figure~\ref{Figure:matchembedding}.}    
    \label{Figure:chessembedding}
\end{figure}

\begin{figure}
\centering
    \framebox{\includegraphics[width=\linewidth, keepaspectratio]{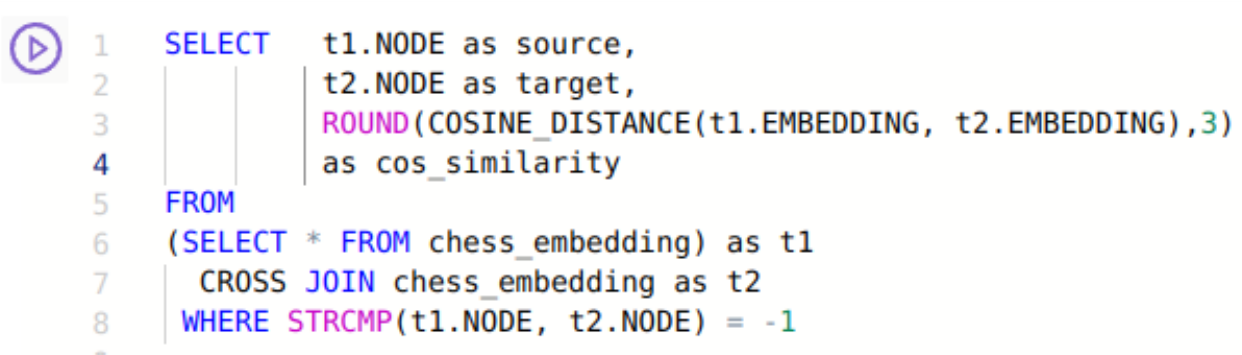}}
    \caption{The Kinetica-SQL statement using a cross join to run the vector similarity analysis between node embeddings of each node pair from the result table, \textit{chess\_embedding} of the embeddding Kinetica-Graph solver.}
    \label{Figure:cossinesimilaritysql}
\end{figure}

The output of the embedding solver is a database table with a vector per graph node as depicted in Figure~\ref{Figure:chessembedding}. These embedding results can be used in any vector similarity functions; such as a cosine similarity as depicted in Figure~\ref{Figure:cossinesimilaritysql}. A common use case for vector similarity is, for instance, in recommendation engines for various industries, from friend recommendations in social networks to the next likely item in your shopping chart. The efficiency and accuracy of these embeddings, however, depends on the rich-ness of the vector sub-features and the sophistication of the randomly selected training sets in optimizing the vector contents. We argue that even the best embedding algorithm would be less accurate compared to the precise connections and labels depicted in the graph topology itself. However, mapping of graphs to vectors has a distinct advantage that they can be applied in a standard manner using simple vector functions in many AI modules. The alternative of using knowledge graph analytics has almost no standardization in many downstream AI applications provided by various graph vendors.

The stipulation of the existing four sub-features representative of the graph topology can certainly be mitigated by either adding more features or a different set of predicates. One other criterion that seems to make sense to include is the distance metric as discussed in~\cite{kgembed1,kgembed2}; which considers similarity for nodes at an equal distance from a set source. However, this statement implies to include all nodes to be similar lying on the same ring-radius distance (hop or weight distance) from the center as the source. This is however a wrong postulate since we know that the nodes on the same ring may be at equal distance away from a source at the center, but they are no-where close to each other particularly for the nodes across each other at any section of the ring. However, along with the cluster index as is already a sub-feature, the combined effect (always consider the inner-product sense) might move the argument to a more acceptable and even preferred state. Another area of future development is in the dynamic additions to the graph, and how to update graph embeddings for the new additions that should be calculable instantly and ready for vector analysis in order for it to be useful in real-time simulations. We are considering to eliminate recomputing of the embeddings for new node insertions by caching and using the results of already computed  weight parameters and interpolating probability and cluster indexes from adjacent nodes instead of running compute heavy cluster and probability solvers. It'll then be up-to the user to decide when to recompute for more accurate embedding values, most probably to be triggered after the number of updates reaching to a significant threshold.

%\begin{figure}
%\centering
%    \framebox{\includegraphics[width=0.4\linewidth, keepaspectratio]{cossimilarity.pdf}}
%    \caption{Cosine similarity results of Figure~\ref{Figure:cossinesimilaritysql} among the node pairs of the \textit{chess} graph.}
%    \label{Figure:cossinesimilarity}
%\end{figure}

\section*{Notes on Contributors}
\small{
\noindent \textbf{Bilge Kaan Karamete} is the lead technologist for the Geospatial and Graph efforts at Kinetica. His research interests include computational geometry/algorithm development, unstructured mesh generation, parallel graph solvers. He holds PhD in Engineering Sciences from the Middle East Technical University, Ankara Turkey, and post doctorate in Computational Sciences from Rensselaer Polytechnic Institute, Troy New York.

\noindent \textbf{Eli Glaser} is VP of Engineering at Kinetica. He leads the development teams concentrating in data analytics, query capability and performance. Eli holds Master's in Electrical Engineering from The Johns Hopkins University, Baltimore Maryland.
}

\section{Software avaliability}

Kinetica and Kinetica-Graph is  freely available in Kinetica's Developer Edition at \textbf{https://www.kinetica.com/try} that the use cases depicted in this manuscript can easily be replicated by the readers.

%\section*{References}

\bibliography{embedding}

\end{document}